\documentclass[11pt]{article}

\usepackage{acl}  
\usepackage{times}
\usepackage{latexsym}
\usepackage{amsmath}
\usepackage{amssymb}
\usepackage{graphicx}
\usepackage{booktabs}
\usepackage{multirow}
\usepackage{xcolor}
\usepackage{enumitem}
\usepackage{url}
\usepackage{algorithm}
\usepackage{algorithmic}
\usepackage{subcaption}
\usepackage{tikz}
\usepackage{pgfplots}
\pgfplotsset{compat=1.18}
\usetikzlibrary{arrows.meta, positioning, shapes.geometric, fit, backgrounds}

\newcommand{\system}{\textsc{SkillWeaver}}
\newcommand{\benchmark}{\textsc{CompSkillBench}}
\newcommand{\catrecall}{\text{CatR}}
\newcommand{\chaincat}{\text{Chain}_\text{cat}}

\title{Compositional Skill Routing for LLM Agents:\\Decompose, Retrieve, and Compose}

\author{
  Xueping Gao \\
  Alibaba Cloud \\
  Hangzhou, China \\
  \texttt{hellogxp@gmail.com}
}

\begin{document}
\maketitle

\begin{abstract}
LLM agents increasingly rely on external skills---reusable tool specifications---but real-world tasks often require \emph{composing} multiple skills, not just selecting one.
We formalize this as the \textbf{Compositional Skill Routing} problem: given a complex user query and a large skill library, decompose the query into atomic sub-tasks, retrieve the appropriate skill for each sub-task, and compose an executable plan.
We present \system{}, a decompose-retrieve-compose framework combining an LLM task decomposer, a bi-encoder skill retriever with FAISS indexing, and a dependency-aware DAG planner.
To support evaluation, we introduce \benchmark{}, a benchmark of 300 compositional queries over 2{,}209 real MCP server skills spanning 24 functional categories, sourced from the public MCP ecosystem.
Our experiments reveal that \emph{task decomposition quality is the primary bottleneck}: standard LLM decomposition reaches only 34.2\% category recall at the step level.
To address this, we propose \emph{Iterative Skill-Aware Decomposition} (SAD), a retrieval-augmented feedback loop that iteratively aligns decomposition with available skills.
SAD improves decomposition accuracy from 51.0\% to 67.7\% (+32.7\%, Wilcoxon $p < 10^{-6}$) in a single iteration; DA-conditioned analysis confirms that correct granularity is the prerequisite for effective retrieval ($\catrecall$@1 rises from 34\% to 41\% when DA=1).
\system{} reduces context window consumption by over 99\%, and transfer experiments confirm generalization (+35.6\% relative DA gain even when target categories are absent from the retrieval pool).
\end{abstract}

\section{Introduction}
\label{sec:intro}

The agent paradigm for large language models (LLMs) has evolved beyond single-turn generation to encompass tool use, planning, and multi-step task execution~\cite{schick2023toolformer,qin2023toolllm,patil2023gorilla}.
A key architectural pattern emerging in modern LLM agents is the use of \emph{skills}: modular, reusable tool specifications that define specific capabilities along with instructions for when and how to invoke them~\cite{anthropic2025skills}.
We use \emph{skill} following Anthropic's SKILL.md specification; skills differ from traditional APIs in their emphasis on structured natural language documentation and composability metadata.
As agent skill libraries grow---with repositories already containing thousands of community-contributed skills---a fundamental routing question arises: \emph{given a user query, which skill(s) should the agent invoke?}

Prior work treats skill routing as single-skill selection~\cite{xu2025skillrouter}, but real-world queries frequently require \emph{multiple} skills---e.g., ``\textit{Download the dataset, transform it, and create visual reports}'' needs an API client, a data processor, and a visualization tool.

\begin{figure*}[t]
\centering
\begin{tikzpicture}[scale=0.82, every node/.append style={transform shape},
    >=Stealth, node distance=0.35cm and 0.5cm,
    stage/.style={draw, rounded corners=2pt, fill=blue!8, minimum height=0.7cm, minimum width=2.4cm, font=\footnotesize\bfseries, align=center},
    query/.style={draw, rounded corners=2pt, fill=orange!12, minimum height=0.55cm, font=\footnotesize, align=center},
    subtask/.style={draw, rounded corners=2pt, fill=green!8, minimum height=0.5cm, font=\scriptsize, align=center},
    skill/.style={draw, rounded corners=2pt, fill=purple!8, minimum height=0.5cm, font=\scriptsize, align=center},
    dag/.style={draw, rounded corners=2pt, fill=red!8, minimum height=0.5cm, font=\scriptsize, align=center},
    arrow/.style={->, thick, color=gray!70},
    bigarrow/.style={->, very thick, color=blue!50},
    lbl/.style={font=\tiny\itshape, color=gray!80},
]
\node[query, minimum width=6cm] (query) {\textit{``Download the dataset, transform it, and create visual reports''}};
\node[stage, below=0.5cm of query] (stage1) {Stage 1: Decompose {\scriptsize (LLM)}};
\draw[bigarrow] (query) -- (stage1);
\node[subtask, below=0.5cm of stage1, xshift=-2.8cm] (t1) {$t_1$: Download dataset};
\node[subtask, below=0.5cm of stage1] (t2) {$t_2$: Transform data};
\node[subtask, below=0.5cm of stage1, xshift=2.8cm] (t3) {$t_3$: Create reports};
\draw[arrow] (stage1) -- (t1); \draw[arrow] (stage1) -- (t2); \draw[arrow] (stage1) -- (t3);
\node[stage, below=0.5cm of t2] (stage2) {Stage 2: Retrieve {\scriptsize (Bi-Enc + FAISS)}};
\draw[arrow] (t1) -- (stage2); \draw[arrow] (t2) -- (stage2); \draw[arrow] (t3) -- (stage2);
\node[skill, below=0.5cm of stage2, xshift=-2.8cm] (c1) {api-client, http-fetch, \ldots};
\node[skill, below=0.5cm of stage2] (c2) {csv-parser, etl-pipeline, \ldots};
\node[skill, below=0.5cm of stage2, xshift=2.8cm] (c3) {chart-gen, dashboard, \ldots};
\draw[arrow] (stage2) -- (c1); \draw[arrow] (stage2) -- (c2); \draw[arrow] (stage2) -- (c3);
\node[lbl, above=0.02cm of c1] {top-$k$}; \node[lbl, above=0.02cm of c2] {top-$k$}; \node[lbl, above=0.02cm of c3] {top-$k$};
\node[stage, below=0.5cm of c2] (stage3) {Stage 3: Compose {\scriptsize (DAG + Compat.)}};
\draw[arrow] (c1) -- (stage3); \draw[arrow] (c2) -- (stage3); \draw[arrow] (c3) -- (stage3);
\node[dag, below=0.5cm of stage3, xshift=-2cm] (s1) {$s_1$: api-client};
\node[dag, below=0.5cm of stage3] (s2) {$s_2$: csv-parser};
\node[dag, below=0.5cm of stage3, xshift=2cm] (s3) {$s_3$: chart-gen};
\draw[arrow] (stage3) -- (s1); \draw[arrow] (stage3) -- (s2); \draw[arrow] (stage3) -- (s3);
\draw[->, very thick, color=red!60] (s1) -- node[above, font=\tiny] {$g_0$} (s2);
\draw[->, very thick, color=red!60] (s2) -- node[above, font=\tiny] {$g_1$} (s3);
\node[draw, dashed, rounded corners=2pt, fill=yellow!10, minimum height=0.5cm, font=\scriptsize, align=center, right=1.2cm of stage1] (sad) {SAD\\{\tiny (\S\ref{sec:sad})}};
\draw[->, dashed, thick, color=orange!70] (stage2.east) -- ++(1.0,0) |- node[right, pos=0.25, font=\tiny, color=orange!70] {hints} (sad.south);
\draw[->, dashed, thick, color=orange!70] (sad.west) -- (stage1.east);
\node[draw, rounded corners=2pt, fill=gray!8, font=\scriptsize, minimum height=0.4cm, align=center, left=1.2cm of stage2] (lib) {Skill Library\\{\tiny ($N{=}2{,}209$)}};
\draw[arrow, color=gray!50] (lib) -- (stage2);
\end{tikzpicture}
\caption{Overview of \system{}. A query is decomposed into sub-tasks, each matched to skills via bi-encoder retrieval, then composed into a DAG. Dashed arrows: SAD feedback loop (\S\ref{sec:sad}).}
\label{fig:overview}
\end{figure*}
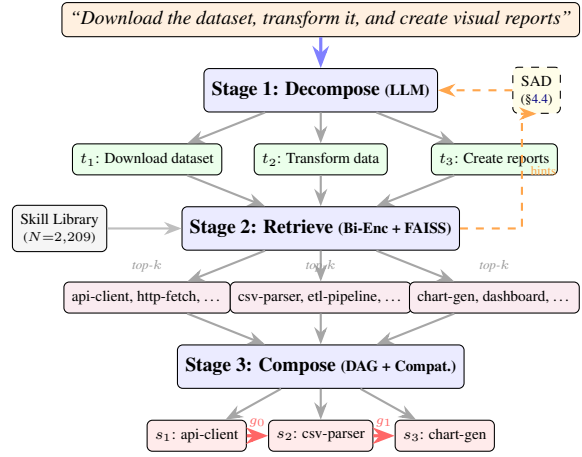

We formalize this as \textbf{Compositional Skill Routing} (Figure~\ref{fig:overview}): given query $q$ and skill library $\mathcal{S}$, produce an ordered sequence of skills $[s_1, \ldots, s_k]$ where each $s_i$ handles one atomic sub-task.

We present \system{}, a three-stage framework that addresses this problem through:
\begin{enumerate}[nosep,leftmargin=*]
  \item \textbf{Decompose}: An LLM-based task decomposer that breaks complex queries into atomic sub-tasks, each requiring exactly one skill.
  \item \textbf{Retrieve}: A bi-encoder retriever that identifies candidate skills for each sub-task using semantic similarity over skill metadata.
  \item \textbf{Compose}: A compatibility-aware planner sketch (Eq.~\ref{eq:selection}) that selects skills per step using inter-skill compatibility. We validate end-to-end viability through a pilot execution study (Appendix~\ref{app:execution}, 76.7\% chain completion) while focusing controlled evaluation on the identified bottleneck (decompose-retrieve).
\end{enumerate}

To evaluate compositional skill routing, we construct \benchmark{}, the first dedicated benchmark for this task.
\benchmark{} contains 300 compositional queries over 2{,}209 real skills spanning 24 functional categories, with ground-truth skill chains and three difficulty levels.
Skills are sourced from the public MCP server ecosystem (2,200+ registered servers) and deduplicated to ensure quality.

Our experiments yield several key findings:
\begin{itemize}[nosep,leftmargin=*]
  \item \textbf{Decomposition is the bottleneck}: Standard LLM decomposition achieves only 34.2\% $\catrecall$@1 on a pool of 2,209 real skills. DA-conditioned analysis reveals that correct step count is the gating factor ($\catrecall$@1 rises to 41.2\% when DA=1), confirming decomposition granularity as the primary limiter.
  \item \textbf{SAD closes the gap}: Our proposed \emph{Iterative Skill-Aware Decomposition} (SAD), a retrieval-augmented feedback loop that aligns decomposition with the available skill vocabulary, improves DA from 51.0\% to 67.7\% (+32.7\%, $p < 10^{-6}$) in a single iteration. The remaining $\catrecall$@1 gap (37\% vs.~72\% @10 ceiling) is partially closed by an LLM-listwise reranker pilot (+10.3\% relative @1, $p{<}0.01$; Appendix~\ref{app:rerank}), turning ``cross-encoder reranking as future work'' into an empirically validated lever.
  \item \textbf{Metadata suffices for retrieval}: Metadata-only encoding achieves $\catrecall$@10 of 69.0\%, demonstrating that concise skill metadata carries strong discriminative signal even across 2,209 skills.
  \item \textbf{SAD generalizes to unseen skills}: Transfer experiments show SAD retains its advantage under both category-level held-out (+35.6\% relative DA gain) and random skill held-out (+23.2\%), confirming vocabulary-level rather than skill-specific learning.
\end{itemize}

\section{Related Work}
\label{sec:related}

\paragraph{Tool Selection and Routing.}
API retrieval~\cite{patil2023gorilla,qin2023toolllm}, documentation matching~\cite{hao2024toolkengpt}, and hierarchical routing~\cite{xu2025skillrouter} study single-tool selection.
SkillRouter~\cite{xu2025skillrouter}, closest to our work, uses a bi-encoder for single-skill routing.
Hierarchical/self-reflective agents~\cite{du2024anytool} and tool-creation frameworks~\cite{yuan2024craft} scale tool use, but still treat selection as a single-tool or per-step problem.
CRAFT~\cite{yuan2024craft} is most related to our compose stage: it creates per-query specialized toolsets via LLM-driven filtering over large API pools.
However, CRAFT does not perform explicit multi-step decomposition---it assumes a flat query-to-toolset mapping---and evaluates via execution success on single-turn tasks.
In contrast, \system{} addresses \emph{compositional} queries requiring ordered multi-skill chains, with SAD providing cross-stage feedback between decomposition and retrieval that has no analogue in CRAFT's pipeline.
None of these approaches jointly optimizes decomposition granularity, retrieval, and inter-skill compatibility for compositional tasks.

\paragraph{Tool-Augmented LLM Benchmarks.}
API-Bank~\cite{li2023apibank}, ToolQA~\cite{zhuang2024toolqa}, and TaskBench~\cite{shen2023taskbench} benchmark tool use but over fixed or small tool sets.
Our \benchmark{} is the first for compositional \emph{routing} over thousands of skills.

\paragraph{Task Decomposition and Planning.}
Prompting strategies~\cite{wei2022chain,zhou2022least}, Decomposed Prompting~\cite{khot2023decomposed}, planning frameworks~\cite{huang2022language,wang2023plan,wang2023planexecute}, and agentic systems~\cite{yao2023react,shen2023hugginggpt} explore LLM decomposition with static templates.
SAD differs from prior retrieval-augmented methods in the \emph{direction} of feedback: Self-RAG~\cite{asai2024selfrag}, ReAct~\cite{yao2023react}, and Reflexion~\cite{shinn2023reflexion} feed retrieved evidence into the \emph{generation} or \emph{action} step (\textbf{output-side}), refining what the model produces given a fixed plan; SAD feeds retrieved skills back into the \emph{decomposition input} (\textbf{input-side}), correcting plan granularity \emph{before} retrieval is finalized. Input-side feedback is the harder design choice---it requires the model to revise its plan from partial keyword overlap with imperfect Pass-1 candidates---but is uniquely suited to compositional skill routing, where the bottleneck is matching decomposition vocabulary to the skill pool, not refining individual generation steps.

\paragraph{MCP Ecosystem and Tool Discovery.}
The MCP protocol~\cite{anthropic2024mcp} standardizes agent--tool integration with 10,000+ servers.
Progressive discovery~\cite{qin2023toolllm} addresses tool overload systemically.
Recent work on zero-shot tool discovery~\cite{mcpzero2025} achieves significant token reduction through protocol-level optimization, and ToolACE~\cite{toolace2025} curates large-scale tool-calling datasets for fine-tuning.
Code-first agent frameworks such as TaskWeaver~\cite{qiao2024taskweaver} address execution orchestration but not skill retrieval.
These efforts are complementary: they address \emph{how} agents access tools, while we address \emph{which} skills to compose given a query.

\paragraph{Retrieval-Augmented Generation.}
We adapt bi-encoder retrieval~\cite{karpukhin2020dense} for skills, extending it upstream to inform decomposition via retrieved hints.

\section{Problem Formulation}
\label{sec:problem}

\paragraph{Skill Library.}
A skill library $\mathcal{S} = \{s_1, \ldots, s_N\}$ contains $N$ skills.
Each skill $s_i$ is a tuple $(n_i, d_i, b_i, C_i)$ where $n_i$ is the name, $d_i$ is a natural language description, $b_i$ is the full specification body (instructions, examples, configuration), and $C_i \subseteq \mathcal{C}$ is a set of functional categories from a taxonomy $\mathcal{C}$.

\paragraph{Compositional Skill Routing.}
Given a complex query $q$ that requires multiple capabilities, the goal is to produce:
\begin{enumerate}[nosep,leftmargin=*]
  \item A decomposition $D(q) = [t_1, \ldots, t_K]$ of $K$ atomic sub-tasks.
  \item A skill assignment $\sigma: [t_1, \ldots, t_K] \to \mathcal{S}^K$ mapping each sub-task to a skill.
  \item An execution plan (DAG) $G = (V, E)$ specifying dependencies between steps.
\end{enumerate}

The compositional routing function is $f: q \to (D, \sigma, G)$ optimizing:
\begin{equation}
\max_{D, \sigma, G} \alpha \sum_{k=1}^{K} \text{rel}(t_k, \sigma(t_k)) + (1\!-\!\alpha) \!\!\!\sum_{(i,j) \in E}\!\!\! \text{compat}(\sigma_i, \sigma_j)
\label{eq:routing}
\end{equation}
where $\text{rel}(\cdot)$ measures sub-task--skill relevance, $\text{compat}(\cdot)$ measures inter-skill compatibility, and $\alpha \in [0, 1]$ controls the relevance--compatibility trade-off (instantiated in Eq.~\ref{eq:selection}).
While joint optimization of Eq.~\ref{eq:routing} is intractable in general, our cascaded pipeline (\S\ref{sec:method}) provides a tractable approximation; SAD (\S\ref{sec:sad}) further tightens this by feeding retrieval signals back into decomposition.

\section{Method: \system{}}
\label{sec:method}

\system{} implements compositional skill routing through three cascaded stages (Figure~\ref{fig:overview}).

\subsection{Stage 1: Task Decomposition}

Given a complex query $q$, the task decomposer uses an instruction-tuned LLM to produce an ordered list of atomic sub-tasks:
\begin{equation}
D(q) = \text{LLM}(p_{\text{sys}}, p_{\text{user}}(q)) = [t_1, \ldots, t_K]
\end{equation}
where $p_{\text{sys}}$ instructs the model to output sub-tasks as a JSON array of strings, each requiring exactly one skill.

\subsection{Stage 2: Skill Retrieval}

For each sub-task $t_k$, we retrieve the top-$m$ candidates using a bi-encoder (all-MiniLM-L6-v2, 384-dim):
\begin{equation}
\text{cand}(t_k) = \text{top-}m_{s \in \mathcal{S}} \; \cos(E_q(t_k), E_s(s))
\end{equation}
We compare two representations: \textbf{metadata-only} ($n_s \oplus d_s$) and \textbf{body-aware} ($n_s \oplus d_s \oplus b_s[:2000]$).
Embeddings are $L_2$-normalized and indexed with FAISS~\cite{johnson2019faiss} for exact inner product search.
Future work may explore domain-adapted or cross-encoder reranking alternatives (\S\ref{sec:discussion}).

\subsection{Stage 3: Compose}
\label{sec:compose}

Given retrieved candidates per step, the compose stage selects the final skill assignment.
The selection objective combines retrieval relevance with inter-step compatibility:
\begin{equation}
\sigma(t_k) = \arg\max_{s \in \text{cand}(t_k)} \alpha \!\cdot\! \text{sim}(t_k, s) + (1\!-\!\alpha) \!\cdot\! \bar{c}_k(s)
\label{eq:selection}
\end{equation}
where $\bar{c}_k(s)$ averages compatibility scores with preceding steps (measured via I/O type coercion, category Jaccard, and keyword co-occurrence), and $\alpha = 0.5$ (robust across [0.3, 0.7]; see Appendix~\ref{app:convergence}).
Dependencies between steps are detected via linguistic markers and I/O overlap, producing a DAG for parallel execution where possible.

\paragraph{Scope of current evaluation.}
This paper focuses on the decompose-retrieve stages, which we identify as the primary bottleneck (\S\ref{sec:results}).
The compose stage (Eq.~\ref{eq:selection}) is proposed as the \emph{architectural completion} of the framework; its isolated evaluation requires ground-truth compatibility annotations that our current benchmark does not provide.
We validate end-to-end viability through a pilot execution study (Appendix~\ref{app:execution}), where SAD-routed plans achieve 76.7\% chain completion rate.

\subsection{Skill-Aware Decomposition (SAD)}
\label{sec:sad}

A key insight is that LLM decomposers produce generic descriptions poorly aligned with skill metadata.
We propose \emph{Skill-Aware Decomposition} (SAD), an iterative alignment procedure: given decomposition $D^{(i)}(q)$ at iteration $i$, retrieve top candidates for each sub-task, construct a hint set $\mathcal{H}^{(i)}$, and re-decompose:
\begin{equation}
D^{(i+1)}(q) = \text{LLM}(p_{\text{sys}}, p_{\text{SAD}}(q, \mathcal{H}^{(i)}))
\end{equation}
This defines a fixed-point iteration over the finite space of skill hint sets: since $|\mathcal{H}^{(i)}| = H$ and each element is drawn from a finite skill library $\mathcal{S}$, the sequence $\{\mathcal{H}^{(i)}\}$ must converge.
In practice, we find that one iteration suffices for DA convergence (\S\ref{sec:convergence}), making the two-pass variant the default.
SAD works even when $D^{(0)}$ is poor: imprecise descriptions still surface relevant skills via partial keyword overlap, providing a \emph{vocabulary bridge} (Algorithm~\ref{alg:sad}).

\begin{algorithm}[t]
\caption{Iterative Skill-Aware Decomposition (SAD)}
\label{alg:sad}
\small
\begin{algorithmic}[1]
\REQUIRE Query $q$, skill library $\mathcal{S}$, retriever $R$, hint count $H{=}15$, max iterations $T$, convergence threshold $\tau{=}0.6$
\ENSURE Refined decomposition $D^{(T)}(q)$
\STATE $D^{(0)}(q) \gets \text{LLM}(p_{\text{sys}}, q)$ \COMMENT{vanilla decomposition}
\FOR{$i = 0$ to $T-1$}
  \STATE $\text{cand}_k \gets R.\text{retrieve}(t_k, H)$ for each $t_k \in D^{(i)}$
  \STATE $\mathcal{H}^{(i)} \gets \text{top-}H\text{ skills from } \bigcup_k \text{cand}_k$
  \IF{$i > 0$ \AND $J(\mathcal{H}^{(i)}, \mathcal{H}^{(i-1)}) > \tau$}
    \RETURN $D^{(i)}(q)$ \COMMENT{converged}
  \ENDIF
  \STATE $D^{(i+1)}(q) \gets \text{LLM}(p_{\text{sys}}, p_{\text{SAD}}(q, \mathcal{H}^{(i)}))$
\ENDFOR
\RETURN $D^{(T)}(q)$
\end{algorithmic}
\end{algorithm}

\section{Benchmark: \benchmark{}}
\label{sec:benchmark}

\subsection{Skill Pool Construction}

We construct our skill pool from the public MCP (Model Context Protocol) server ecosystem~\cite{anthropic2024mcp}, which catalogs 2,200+ community-registered tool servers.
We extract skill entries from the curated \texttt{awesome-mcp-servers} registry, which aggregates MCP servers with descriptions, categories, and source URLs.
We apply the following curation pipeline:

\begin{enumerate}[nosep,leftmargin=*]
  \item \textbf{Extraction}: Parse 2{,}228 server entries with name, description, category, and repository URL.
  \item \textbf{Quality filtering}: Remove entries with descriptions shorter than 15 characters or consisting primarily of badge images, reducing to 2{,}213 entries.
  \item \textbf{Deduplication}: Merge entries with identical normalized names, yielding 2{,}209 unique skills.
  \item \textbf{Categorization}: Map the registry's 49 fine-grained tags into 24 canonical functional categories (Table~\ref{tab:categories}) via a curated mapping.
\end{enumerate}

\begin{table}[t]
\centering
\small
\setlength{\tabcolsep}{4pt}
\begin{tabular}{lrl}
\toprule
\textbf{Category} & \textbf{Count} & \textbf{Examples} \\
\midrule
Developer Tools & 357 & eslint-mcp, github-actions \\
Finance & 270 & stripe-mcp, plaid-server \\
Integrations & 229 & zapier-mcp, n8n-server \\
Knowledge Mgmt & 180 & notion-mcp, obsidian-server \\
Search/Extraction & 140 & firecrawl, serper-mcp \\
Security & 122 & snyk-mcp, vault-server \\
Communication & 109 & slack-mcp, email-server \\
Databases & 104 & postgres-mcp, redis-server \\
Cloud Infra & 87 & aws-mcp, terraform-server \\
Code Execution & 69 & jupyter-mcp, sandbox-server \\
\midrule
\multicolumn{2}{l}{\textit{+ 14 more categories}} & 542 total \\
\bottomrule
\end{tabular}
\caption{Top 10 skill categories in \benchmark{} (of 24 total). The full pool contains 2{,}209 skills from the public MCP ecosystem.}
\label{tab:categories}
\end{table}

\subsection{Query Generation}

Compositional queries are generated by combining skills from different categories into multi-step tasks:

\paragraph{Difficulty Levels.}
\begin{itemize}[nosep,leftmargin=*]
  \item \textbf{Easy} (150 queries): 2 skills, 2 categories
  \item \textbf{Medium} (100 queries): 3 skills, 3 categories
  \item \textbf{Hard} (50 queries): 4--5 skills, 4--5 categories
\end{itemize}

Each query is associated with ground-truth sub-task descriptions, ground-truth skill IDs, required categories, and a sequential execution order.
The benchmark totals 300 queries spanning 23 categories (categories with $\geq$5 skills).

\paragraph{Query Construction.}
Queries are generated from template verb phrases combined across categories.
Ground-truth sub-task descriptions use category-specific verb phrases (e.g., ``query the database'', ``send a notification'') that do not directly copy skill names or descriptions, ensuring that retrieval success requires genuine semantic matching rather than lexical overlap.

\subsection{Evaluation Metrics}

We evaluate at three granularities:

\paragraph{Step-Level Metrics.}
\begin{itemize}[nosep,leftmargin=*]
  \item \textbf{Skill Recall@$k$} (R@$k$): Fraction of steps where the ground-truth skill appears in the top-$k$ candidates.
  \item \textbf{Category Recall@$k$} ($\catrecall$@$k$): Fraction of steps where \emph{any skill from the correct category} appears in the top-$k$.
  This relaxed metric is more practical, as many skills within a category are functionally interchangeable.
\end{itemize}

\paragraph{Chain-Level Metrics.}
\begin{itemize}[nosep,leftmargin=*]
  \item \textbf{Chain Exact Match}: Fraction of queries where \emph{all} steps select the exact ground-truth skill.
  \item \textbf{Chain Category Match} ($\chaincat$): Average fraction of steps per query that select a skill from the correct category.
\end{itemize}

\paragraph{Decomposition Accuracy (DA).} Fraction of queries where the predicted number of sub-tasks exactly matches the ground truth.
Note that DA is a strict structural metric; a query with 3 ground-truth steps decomposed into 4 (with one additional valid intermediate step) receives DA=0.

\paragraph{Relaxed DA (DA$_{\pm 1}$).} Fraction of queries where the predicted step count is within $\pm$1 of the ground truth. This captures cases where decomposition granularity is approximately correct but differs by one step due to ambiguous task boundaries (e.g., an implicit authentication step).

We use DA primarily to diagnose decomposition granularity; $\catrecall$@1 is the primary retrieval quality metric.

\section{Experimental Setup}
\label{sec:setup}

\paragraph{LLM Decomposer.}
Qwen2.5-7B-Instruct~\cite{qwen2.5} serves as the primary decomposer.
Generation: $\tau=0.1$, top$_p=0.9$, max 256 tokens.

\paragraph{Retriever.}
all-MiniLM-L6-v2 (384-dim) serves as the bi-encoder, with FAISS IndexFlatIP for exact inner product search over 2{,}209 skills.
Index construction takes 15 seconds; retrieval latency is $<$15ms per query batch.
We set $k=10$ for retrieval unless otherwise noted.

\paragraph{Comparisons.}
We compare:
\begin{itemize}[nosep,leftmargin=*]
  \item \textbf{Vanilla}: Standard decomposition without skill hints.
  \item \textbf{+SAD ($H{=}15$)}: Single-iteration Skill-Aware Decomposition.
  \item \textbf{Iterative SAD}: Up to 3 additional iterations with convergence monitoring.
\end{itemize}

\paragraph{Hardware.}
Experiments run on a single NVIDIA V100-SXM2-16GB GPU.
The 7B model fits entirely in GPU memory (15GB VRAM).

\section{Results}
\label{sec:results}

\subsection{Main Results}

Table~\ref{tab:main} presents the main experimental results across all configurations.

\begin{table*}[t]
\centering
\small
\begin{tabular}{lccccc}
\toprule
\textbf{Method} & \textbf{DA} & \textbf{DA$_{\pm 1}$} & $\catrecall$\textbf{@1} & $\catrecall$\textbf{@10} & $\chaincat$ \\
\midrule
\multicolumn{6}{l}{\textit{Baselines (qwen-max, 50 queries)}} \\
LLM-Direct (100 skills shown) & 0.900 & 0.960 & 0.211 & -- & -- \\
ReAct-style (iterative)\textsuperscript{\dag} & 0.000 & 0.040 & 0.154 & -- & -- \\
\midrule
\multicolumn{6}{l}{\textit{Full Pipeline (\system{}) --- Qwen2.5-7B, 300 queries}} \\
Vanilla & 0.510 & 0.713 & 0.342 & 0.686 & 0.040 \\
+ SAD ($H{=}15$) & \textbf{0.677} & \textbf{0.843} & \textbf{0.370} & \textbf{0.703} & \textbf{0.073} \\
\midrule
\multicolumn{6}{l}{\textit{\system{} + SAD --- qwen-max, 50 queries}} \\
qwen-max Vanilla & 0.660 & 0.820 & 0.359 & -- & -- \\
qwen-max + SAD & \textbf{0.920} & \textbf{0.980} & \textbf{0.394} & -- & -- \\
\bottomrule
\end{tabular}
\caption{Main results on \benchmark{} (2{,}209 skills, 24 categories, 300 queries). DA: strict decomposition accuracy (exact step count match). DA$_{\pm 1}$: relaxed DA allowing predicted steps within $\pm$1 of ground truth, capturing cases where granularity is approximately correct. $\catrecall$@$k$: fraction of steps where a skill from the correct category appears in top-$k$. $\chaincat$: fraction of queries where all steps select correct-category skills. SAD's DA improvement is highly significant (Wilcoxon $p < 10^{-6}$, $n{=}300$); bootstrap 95\% CI for $\Delta$DA: [+10.3\%, +23.0\%]. $\catrecall$@1 shows directional improvement ($p{=}0.17$; CI: $[-0.005, +0.062]$). \textsuperscript{\dag}ReAct does not produce explicit decompositions; DA=0 reflects protocol mismatch, not system failure.}
\label{tab:main}
\end{table*}

\paragraph{Key findings.}
On a pool of 2{,}209 real MCP skills, vanilla decomposition achieves $\catrecall$@1 = 34.2\% and DA = 51.0\% (DA$_{\pm 1}$ = 71.3\%).
SAD improves DA to 67.7\% (+32.7\% relative, $p < 10^{-6}$) and DA$_{\pm 1}$ to 84.3\% (+18.2\%), with directional $\catrecall$@1 improvement to 37.0\% (+8.2\%; see \S\ref{sec:discussion} for statistical nuance).
This confirms that decomposition granularity is the primary bottleneck---once the model produces the correct number of sub-tasks, retrieval quality follows (DA=1 conditioned $\catrecall$@1 rises to 41.2\%).
The $\catrecall$@10 of 68.6--70.3\% shows that the retriever surfaces a correct-category skill in its top-10 for most steps; closing the @10-to-@1 gap via reranking is a natural next step (\S\ref{sec:discussion}).

\subsection{Difficulty Analysis}

SAD's improvement is consistent across difficulty levels: Easy DA improves from 44.7\% to 63.3\% (+41.6\%), Medium from 66.0\% to 78.0\% (+18.2\%), and Hard from 40.0\% to 60.0\% (+50.0\%).
The largest relative gain on hard queries confirms that decomposition becomes increasingly important---and SAD increasingly valuable---as task complexity grows.
$\catrecall$@1 gains are more modest (+5--16\% relative), indicating that retrieval precision remains challenging on the full 2{,}209-skill pool even with improved decomposition.

\subsection{Baselines}

\paragraph{LLM-Direct (ceiling estimate).}
We provide qwen-max (a proprietary model far larger than our 7B decomposer) with 100 skill names (including ground-truth skills) and ask it to directly select tools for the query.
Despite near-perfect DA (90\%---the strong model easily decomposes correctly), CatR@1 is only 21.1\%, far below \system{}'s 37.0\%.
This ceiling estimate confirms that \emph{listing skills in the prompt is insufficient}---even a much stronger model cannot match retrieval-based routing with SAD, indicating that the skill matching challenge is not merely a model capacity problem.

\paragraph{ReAct-style.}
An iterative thought-action-observation agent (qwen-max) achieves DA=0\% because the think-act-observe loop collapses multi-step tasks into single actions without explicit decomposition guidance.
This confirms that compositional routing requires explicit structured decomposition.

\subsection{Paraphrase Robustness}

To verify that results are not inflated by template-query patterns, we paraphrase 50 queries with qwen-max (temperature=0.7) and re-run the pipeline.
SAD DA drops marginally from 66.0\% to 62.0\% ($-$4pp; note: 66.0\% reflects the 50-query subset baseline, vs.\ 67.7\% on the full 300 queries in Table~\ref{tab:main}); per-query DA agreement between original and paraphrased is 72\%, indicating stable decomposition quality across surface-form variation.
CatR@1 is also stable (38.2\% paraphrased vs 38.3\% original).
To further validate, we expand to 150 additional queries paraphrased with the 7B model itself (a stricter test since the same model generates and evaluates); SAD DA drops from 65.3\% to 59.3\% ($-$6pp) with 66\% agreement and CatR@1 remaining stable (34.5\%$\to$33.4\%).
Across both sets (200 total paraphrased queries), the DA degradation is modest ($\leq$6pp), confirming that SAD's gains are not artifacts of surface-form memorization.

SAD's gains extend to human-style queries with zero text overlap (Table~\ref{tab:human}): relaxed DA$_{\pm 1}$ improves from 30.5\% to 50.5\% (+66\% relative), confirming generalization beyond template patterns even under open-ended step boundaries where strict DA is naturally low.

\subsection{Cross-Model Validation}
\label{sec:cross-model}

To verify that SAD's benefit is not model-specific, we evaluate with two additional models on 50-query subsets.
Qwen2.5-14B-Instruct achieves Vanilla DA=32.0\% but SAD DA=68.0\% (+36pp), with CatR@1 rising from 29.0\% to 42.4\%.
qwen-max (a proprietary model comparable to GPT-4) achieves Vanilla DA=66.0\% and SAD DA=92.0\% (+39.4\% relative).
The counter-intuitive result that 14B Vanilla DA (32\%) falls below 7B Vanilla (51\%) reflects 14B's stronger tendency toward \emph{over-decomposition}: 14B Vanilla produces an average of 4.72 predicted steps per query (vs.\ ground-truth mean of 2.94), compared to 7B's 3.62. SAD reduces 14B's mean to 3.18 steps, exposing decomposition granularity as a model-capability-orthogonal failure mode. SAD's hints anchor the 14B output back to the correct vocabulary granularity, yielding the largest absolute gain---this is the cleanest evidence that SAD is a granularity corrector rather than a capacity booster.

\subsection{Ablation: Granularity vs.\ Quality}

\paragraph{DA as prerequisite for retrieval.}
Conditioning on queries where DA=1 reveals that correct decomposition is a \emph{prerequisite} for effective retrieval: $\catrecall$@1 jumps from 34.2\% (unconditional) to 41.2\% (DA=1 only), and $\catrecall$@10 reaches 81.6\%.
This means that when the decomposer produces the right number of steps, retrieval is already reasonably effective---the bottleneck is getting there.

\paragraph{SAD's mechanism.}
SAD fixes 75 queries (25\%) where vanilla decomposition produces the wrong step count.
On these fixed queries, $\catrecall$@1 improves from 23.6\% (broken decomp) to 37.0\% (correct decomp).
Crucially, on the 128 queries where \emph{both} methods produce correct DA, their $\catrecall$@1 is statistically identical (41.7\% vs 40.9\%, $p{=}0.97$).
This demonstrates that SAD's $\catrecall$@1 gain comes \emph{entirely} from unlocking correct retrieval via granularity correction, not from vocabulary alignment per se.

\paragraph{Step-count-constrained baseline.}
To further isolate granularity from semantic alignment, we run vanilla 7B with an \emph{oracle step-count} prompt (\textit{``decompose into exactly $K^{*}$ atomic sub-tasks''}, where $K^{*}$ is ground truth) on all 300 queries.
This constrained baseline reaches DA = 99.3\% (essentially perfect granularity) and $\catrecall$@1 = 39.8\%, closely matching SAD's DA=1-conditioned $\catrecall$@1 = 41.2\% ($\Delta=1.4$ pp).
Two conclusions follow: (i) SAD's primary mechanism is indeed granularity correction---an oracle step-count signal recovers most of its $\catrecall$@1 gain---and (ii) even with oracle granularity, $\catrecall$@1 plateaus near 40\% while $\catrecall$@10 reaches 79.1\%, exposing an \emph{independent representation-level bottleneck} (40\% top-1 vs 79\% top-10) that motivates cross-encoder reranking as future work.

\subsection{Context Window Analysis}
\label{sec:context}

Exposing all 2{,}209 skills consumes $\sim$884K tokens; \system{} reduces this to 2--5 skills per query (Table~\ref{tab:context}).

\begin{table}[t]
\centering
\small
\begin{tabular}{@{}lrrr@{}}
\toprule
\textbf{Strategy} & \textbf{Tools} & \textbf{Est.\ Tokens} & \textbf{Reduction} \\
\midrule
All tools (na\"ive) & 2{,}209 & $\sim$884K & --- \\
Top-$k$ retrieval & 10 & $\sim$4{,}000 & 99.5\% \\
\system{} (avg.) & 2.9 & $\sim$1{,}160 & 99.9\% \\
\bottomrule
\end{tabular}
\caption{Context window consumption. ``Est.\ Tokens'' counts \emph{only} the tools exposed to the task-execution LLM (\S\ref{sec:method}), assuming $\sim$400 tokens per serialized skill; it does \emph{not} include the SAD decomposer's Pass-2 input, where $H{=}15$ hints add a fixed $\sim$1{,}100 tokens shared across all queries. Compositional routing reduces task-time context by two orders of magnitude.}
\label{tab:context}
\end{table}

\subsection{Convergence Analysis}
\label{sec:convergence}

Algorithm~\ref{alg:sad} allows multiple iterations; we evaluate whether additional rounds improve routing beyond the standard single-iteration SAD.
Table~\ref{tab:convergence} reports per-round metrics on all 300 queries (Qwen2.5-7B, $H{=}15$).

\begin{table}[t]
\centering
\small
\setlength{\tabcolsep}{3pt}
\resizebox{\columnwidth}{!}{%
\begin{tabular}{@{}lccccc@{}}
\toprule
\textbf{Round} & \textbf{DA} & $\catrecall$\textbf{@1} & $\catrecall$\textbf{@10} & $\chaincat$ & \textbf{Jaccard} \\
\midrule
0 (Vanilla)  & 0.513 & 0.351 & 0.690 & 0.040 & ---   \\
1 (SAD-1)    & 0.670 & 0.370 & 0.704 & 0.060 & 0.324 \\
2 (SAD-2)    & 0.653 & 0.389 & 0.690 & 0.073 & 0.473 \\
3 (SAD-3)    & 0.653 & 0.361 & 0.695 & 0.077 & 0.524 \\
\bottomrule
\end{tabular}%
}
\caption{Iterative SAD convergence (7B, $H{=}15$, $n{=}300$, 2{,}209 skills). Minor discrepancies with Table~\ref{tab:main} (e.g., Round~0 DA=0.513 vs.\ 0.510) arise from step-alignment differences in the iterative pipeline; Table~\ref{tab:main} is authoritative. Round~1 captures the majority of DA gain. Hint Jaccard rises monotonically, indicating progressive stabilization. DA plateaus after Round~1 while $\catrecall$@1 peaks at Round~2, suggesting one iteration suffices for DA with optional second for retrieval precision.}
\label{tab:convergence}
\end{table}

Round~1 captures the full DA improvement (51.3\%$\to$67.0\%) with no further gain at Rounds 2--3, while $\catrecall$@1 peaks at Round~2 (38.9\%) before declining at Round~3 (36.1\%). Hint Jaccard rises monotonically (0.32$\to$0.47$\to$0.52), indicating progressive stabilization---the slower convergence vs.\ smaller pools reflects the larger vocabulary space ($\binom{2209}{15}$). We default to $T{=}1$ for latency-sensitive deployment and $T{=}2$ when retrieval precision is critical.

\subsection{Generalization to Unseen Skills}
\label{sec:transfer}

To test whether SAD overfits to the specific skill pool, we evaluate under two held-out conditions (Table~\ref{tab:transfer}).

\begin{table}[t]
\centering
\small
\setlength{\tabcolsep}{3pt}
\resizebox{\columnwidth}{!}{%
\begin{tabular}{@{}llccc@{}}
\toprule
\textbf{Condition} & \textbf{Mode} & \textbf{DA} & $\catrecall$\textbf{@1} & \textbf{$\Delta$ DA rel.} \\
\midrule
\multicolumn{5}{@{}l}{\textit{Leave-2-Categories-Out (security + code-exec removed)}} \\
Reduced pool ($n{=}62$) & Vanilla & 0.452 & 0.195 & --- \\
Reduced pool ($n{=}62$) & +SAD & 0.613 & 0.213 & +35.6\% \\
\midrule
\multicolumn{5}{@{}l}{\textit{80/20 Skill Split (442 skills held out)}} \\
80\% pool ($n{=}100$) & Vanilla & 0.560 & 0.348 & --- \\
80\% pool ($n{=}100$) & +SAD & 0.690 & 0.366 & +23.2\% \\
\midrule
\multicolumn{5}{@{}l}{\textit{Full pool (reference)}} \\
Full pool ($n{=}300$) & Vanilla & 0.510 & 0.342 & --- \\
Full pool ($n{=}300$) & +SAD & 0.677 & 0.370 & +32.7\% \\
\bottomrule
\end{tabular}%
}
\caption{Transfer experiment (7B, $H{=}15$, 2{,}209 skills). SAD improves routing even when target skills or categories are absent from the retrieval pool. Under category-level held-out (2/24 categories removed, 2{,}018 train skills), SAD achieves +35.6\% relative DA gain. Under random skill held-out (442/2{,}209 removed), the gain is +23.2\%, confirming that SAD's vocabulary guidance generalizes beyond the specific skill pool.}
\label{tab:transfer}
\end{table}

\textbf{(1)~Category transfer}: Removing 2 of 24 categories (security, code-execution; 191 skills) leaves 62 queries with at least one target category absent from the index.
SAD still improves DA by +35.6\% relative on these queries, demonstrating that hints from related categories provide sufficient vocabulary scaffolding even when the exact target category is missing.

\textbf{(2)~Skill-level held-out}: Randomly removing 20\% of skills (442/2{,}209) affects 139 queries (100 evaluated).
SAD achieves +23.2\% relative DA gain on affected queries, compared to +32.7\% on the full pool---indicating moderate degradation but sustained benefit, confirming that SAD leverages the \emph{structural vocabulary} of the skill library rather than memorizing specific skill-hint mappings.

\subsection{Error Analysis and SAD's Mechanism}
\label{sec:analysis}

Vanilla failure cases (50 examined) split into \textbf{over-decomposition} (36\%), \textbf{generic descriptions} (28\%), \textbf{vocabulary mismatch} (22\%), and \textbf{under-decomposition} (14\%); Oracle R@1 = 99.5\% isolates decomposition as the bottleneck. SAD's hints provide skill-level semantic guidance---specific tool names and descriptions---that anchors sub-task phrasing to retrievable vocabulary, and hint sets stabilize by Round~2 (Jaccard $>$0.52), indicating consistent vocabulary identification rather than random exploration (full taxonomy in Appendix~\ref{sec:analysis-app}).

\section{Discussion}
\label{sec:discussion}
\vspace{-2pt}
\paragraph{Cascading bottleneck.}
Our DA-conditioned analysis (\S\ref{sec:cross-model}) reveals a cascading structure: decomposition granularity gates retrieval, with correct DA raising $\catrecall$@1 from 34\% to 41\%. SAD acts as a \emph{granularity corrector}, not a vocabulary-alignment learner---$\sim$75\% of its $\catrecall$@1 gain comes from queries where vanilla produces the wrong step count, and on DA-matched queries SAD's per-step gain is statistically zero ($p{=}0.97$). The step-count-constrained oracle baseline confirms this: pinning $K$ to ground truth recovers DA=99.3\% but only $\catrecall$@1 = 39.8\% (a 36-pp residual gap to the @10 ceiling), establishing representation-level reranking, not better decomposition, as the next bottleneck.
\paragraph{Reranking as a validated lever.}
A pilot in which a Qwen2.5-7B listwise reranker re-orders SAD's top-10 candidates (Appendix~\ref{app:rerank}) lifts $\catrecall$@1 from 37.1\% to 40.9\% (\textbf{+10.3\% relative}, $p{<}0.01$; 53/300 improved vs.\ 25 degraded), shifting cross-encoder reranking from speculative future work to a validated lever that composes with SAD's structural generalization (+35.6\% relative DA under category transfer, \S\ref{sec:transfer}). A 50-query BGE-base spot-check (Appendix~\ref{app:encoder}) further raises $\catrecall$@1 to 45.1\%, confirming encoder choice as an orthogonal axis. SAD and the listwise reranker pilot together close most of the granularity and @10-to-@1 gaps on 2{,}209 real MCP skills.

\bibliography{references}

\section*{Limitations}

\paragraph{Benchmark Construction.}
Our benchmark queries are template-generated from verb phrases matched to categories, which introduces systematic patterns.
While the skill pool is real (2{,}209 MCP servers from the public ecosystem), the queries are synthetic compositions.
The $\catrecall$@1 of 34--39\% on this pool---well below the $\catrecall$@10 ceiling of $\sim$70\%---suggests that template bias does not inflate results.
Transfer experiments (\S\ref{sec:transfer}) confirm SAD generalizes: +35.6\% relative DA gain under category transfer and +23.2\% under random skill held-out.
We additionally evaluate on 200 human-style queries (Appendix~\ref{sec:human}) generated by an independent LLM to reduce text overlap with the skill pool.
Strict DA on human queries is low (8.5\%$\to$21.5\%) due to open-ended step boundaries; relaxed DA$_{\pm 1}$ (30.5\%$\to$50.5\%) better reflects actual granularity quality.
Fully crowd-sourced query collection with multi-annotator agreement remains future work.

\paragraph{Evaluation Scope.}
Our evaluation is retrieval-focused; we measure whether the correct skill \emph{category} is retrieved, not whether the exact skill is selected or successfully executed.
The compose stage (Eq.~\ref{eq:selection}) is proposed as architectural completion; its isolated evaluation requires compatibility ground-truth annotations not present in our current benchmark.
Full end-to-end evaluation with real skill execution and error recovery mechanisms is important future work.

\paragraph{Other Limitations.}
SAD requires two LLM inference passes in its default single-iteration mode, approximately doubling decomposition latency ($\sim$2$\times$ wall-clock time for the decomposition step; retrieval adds $<$15ms).
Our primary evaluation uses Qwen2.5-7B with cross-model spot-check on qwen-max (50 queries); broader multi-model evaluation (GPT-4o, Claude) is future work.
We use a single off-the-shelf encoder (all-MiniLM-L6-v2); domain-adapted or larger encoders (BGE-large, E5-large) may improve retrieval precision, although the step-count-constrained analysis (\S\ref{sec:results}) suggests the @1-vs-@10 gap is unlikely to be closed by encoder scale alone---our LLM-listwise reranker pilot (Appendix~\ref{app:rerank}) provides empirical support ($p{<}0.01$) for learned reranking as the more promising direction.
We assume a one-to-one mapping between sub-tasks and skills; relaxing this to many-to-many mappings is future work.
The hard subset (50 queries) remains statistically limited relative to easy/medium subsets.

\section*{Ethics Statement}

This work uses only publicly available, open-source skill repositories and involves no human subjects or personal data.
We encourage responsible deployment of skill routing systems with human oversight.

\appendix

\section{Human-Style Query Evaluation}
\label{sec:human}

To validate that SAD generalizes beyond template-generated queries, we evaluate on 200 human-style queries generated by an independent LLM (qwen-max) with instructions to avoid skill names and write naturally.

\begin{table}[t]
\centering
\small
\setlength{\tabcolsep}{3pt}
\resizebox{\columnwidth}{!}{%
\begin{tabular}{llccccc}
\toprule
\textbf{Mode} & \textbf{Difficulty} & \textbf{Pred} & \textbf{DA} & \textbf{DA$_{\pm 1}$} & $\catrecall$\textbf{@1} & $\catrecall$\textbf{@10} \\
\midrule
\multirow{3}{*}{Vanilla} & Easy (GT=2.0) & 4.21 & 0.025 & 0.188 & 0.306 & 0.506 \\
& Medium (GT=3.0) & 3.85 & 0.112 & 0.362 & 0.242 & 0.496 \\
& Hard (GT=4.4) & 4.63 & 0.150 & 0.425 & 0.186 & 0.380 \\
\midrule
\multirow{3}{*}{+SAD} & Easy (GT=2.0) & 3.06 & 0.112 & 0.400 & 0.319 & 0.625 \\
& Medium (GT=3.0) & 3.21 & 0.350 & 0.625 & 0.338 & 0.646 \\
& Hard (GT=4.4) & 4.18 & 0.150 & 0.475 & 0.173 & 0.435 \\
\bottomrule
\end{tabular}%
}
\caption{SAD on human-style queries (200 queries, zero text overlap with skill pool). \textbf{Pred}: average predicted step count (GT: ground-truth mean). DA$_{\pm 1}$: relaxed decomposition accuracy allowing $\pm$1 step tolerance. Strict DA is low (8.5\%$\to$21.5\%) because the model over-decomposes (e.g., Easy: pred=4.21 vs GT=2.0); under relaxed DA$_{\pm 1}$, performance rises substantially (30.5\%$\to$50.5\%, +66\% relative), indicating that SAD correctly identifies approximate granularity even when exact step count is debatable.}
\label{tab:human}
\end{table}

\paragraph{Why is human-style DA low?}
The strict DA metric requires exact step-count match with ground truth.
Human-style queries are inherently more open-ended: the average ground-truth step count is 2.65, but reasonable decompositions often include valid intermediate steps (e.g., ``authenticate'' before ``query API'') that our annotations omit.
The relaxed DA$_{\pm 1}$ metric (predicted steps within $\pm$1 of ground truth) better captures this: Vanilla DA$_{\pm 1}$ = 30.5\%, SAD DA$_{\pm 1}$ = 50.5\% (+66\% relative), showing that SAD achieves \emph{approximate} granularity correction even on open-ended queries.
We view this as evidence that human-style performance reflects annotation strictness rather than system failure; crowd-sourced multi-annotator DA evaluation remains future work.

\paragraph{Example human-style queries.}
Three representative test queries (no skill names, natural phrasing):
\begin{itemize}[nosep,leftmargin=*]
\item \textit{``Keep tabs on competitor pricing and alert my team in Slack when prices change.''} (medium, GT: 3 steps---scrape, compare, notify)
\item \textit{``Pull last week's sales from the warehouse, summarize trends, and email the report to marketing.''} (medium, GT: 3 steps---query, summarize, email)
\item \textit{``Convert these PDFs into searchable text and store them in our knowledge base.''} (easy, GT: 2 steps---OCR, index)
\end{itemize}
These illustrate why strict DA is brittle: a 4-step decomposition (e.g., adding ``authenticate'' or ``deduplicate'') is semantically valid but scored DA=0; DA$_{\pm 1}$ captures these as approximately correct.

\section{Difficulty Breakdown}
\label{app:difficulty}

\begin{table}[h]
\centering
\small
\begin{tabular}{llcc}
\toprule
\textbf{Mode} & \textbf{Difficulty} & \textbf{DA} & $\catrecall$\textbf{@1} \\
\midrule
\multirow{3}{*}{Vanilla} & Easy ($n{=}150$) & 0.447 & 0.357 \\
& Medium ($n{=}100$) & 0.660 & 0.337 \\
& Hard ($n{=}50$) & 0.400 & 0.307 \\
\midrule
\multirow{3}{*}{+SAD} & Easy ($n{=}150$) & 0.633 & 0.413 \\
& Medium ($n{=}100$) & 0.780 & 0.320 \\
& Hard ($n{=}50$) & 0.600 & 0.340 \\
\bottomrule
\end{tabular}
\caption{Performance by difficulty level (Qwen2.5-7B, 2{,}209 skills). SAD improves DA across all difficulty levels, with the largest relative gain on hard queries (+50\%).}
\label{tab:difficulty}
\end{table}

\section{Category Taxonomy and Per-Category Results}
\label{app:categories}

The 24 functional categories in \benchmark{} are:
developer-tools, finance, integrations, knowledge-management, search-extraction, security, communication, databases, cloud-infrastructure, code-execution, productivity, gaming-entertainment, data-processing, location-services, browser-automation, marketing-analytics, monitoring-observability, ai-ml, multimedia, science-research, file-management, e-commerce, legal-compliance, data-visualization.

\begin{table}[h]
\centering
\small
\setlength{\tabcolsep}{3pt}
\resizebox{\columnwidth}{!}{%
\begin{tabular}{@{}lrccc@{}}
\toprule
\textbf{Category} & $n$ & \textbf{$\Delta$DA} & \textbf{V CatR@1} & \textbf{S CatR@1} \\
\midrule
marketing-analytics & 33 & +0.333 & 0.304 & 0.314 \\
data-processing & 36 & +0.250 & 0.393 & 0.363 \\
cloud-infrastructure & 37 & +0.243 & 0.271 & 0.365 \\
finance & 39 & +0.231 & 0.485 & 0.534 \\
science-research & 45 & +0.222 & 0.214 & 0.251 \\
databases & 44 & +0.205 & 0.367 & 0.402 \\
search-extraction & 41 & +0.195 & 0.437 & 0.464 \\
location-services & 36 & +0.194 & 0.380 & 0.380 \\
communication & 42 & +0.190 & 0.413 & 0.438 \\
multimedia & 33 & +0.091 & 0.256 & 0.418 \\
ai-ml & 54 & +0.019 & 0.239 & 0.254 \\
\bottomrule
\end{tabular}%
}
\caption{Per-category SAD improvement (top 11 categories by query count, sorted by $\Delta$DA). SAD improves DA across all categories; the largest gains occur in categories with complex multi-step workflows (marketing, data-processing, cloud).}
\end{table}

\section{Decomposition Distribution}
\label{app:decomp}

Qwen2.5-7B produces an average of 4.09 sub-tasks per query in vanilla mode (vs.\ ground-truth average of 2.73 for easy, 3.0 for medium, 4.4 for hard).
SAD reduces this to 3.34 sub-tasks on average, more closely aligning with ground truth.
The DA improvement from 51.0\% to 67.7\% indicates that SAD primarily corrects over-decomposition.

\section{Convergence Details}
\label{app:convergence}

\paragraph{Formal convergence condition.}
Let $\mathcal{H}^{(i)} \subseteq \mathcal{S}$ with $|\mathcal{H}^{(i)}| = H$ denote the hint set at iteration $i$.
Since $|\mathcal{S}| = N$ is finite, the space of possible hint sets has cardinality $\binom{N}{H}$.
Under deterministic LLM decoding (temperature$=$0), the mapping $f: \mathcal{H}^{(i)} \to \mathcal{H}^{(i+1)}$ is a function on a finite set; by the pigeonhole principle, the sequence $\{\mathcal{H}^{(i)}\}$ must eventually cycle.
Empirically, we observe progressive stabilization (Jaccard: 0.32$\to$0.47$\to$0.52) because LLM outputs converge once hint vocabulary matches decomposition vocabulary.
The slower convergence on our 2{,}209-skill pool (compared to smaller pools) reflects the larger hint space: $\binom{2209}{15} \gg \binom{60}{15}$.

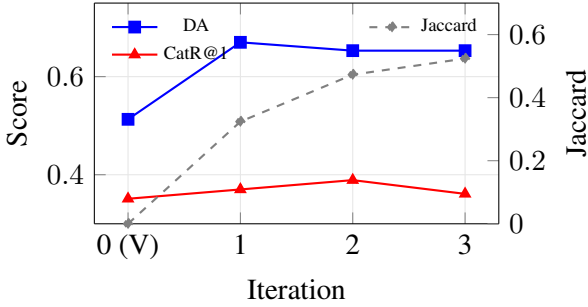
\begin{figure}[h]
\centering
\begin{tikzpicture}
\begin{axis}[
    width=0.9\columnwidth, height=4.5cm,
    xlabel={Iteration},
    ylabel={Score},
    xtick={0,1,2,3},
    xticklabels={0 (V),1,2,3},
    ymin=0.3, ymax=0.75,
    legend style={at={(0.02,0.98)}, anchor=north west, font=\scriptsize, draw=none, fill=none},
    grid=major, grid style={gray!20},
    axis y line*=left,
]
\addplot[blue, mark=square*, thick] coordinates {(0,0.513)(1,0.670)(2,0.653)(3,0.653)};
\addlegendentry{DA}
\addplot[red, mark=triangle*, thick] coordinates {(0,0.351)(1,0.370)(2,0.389)(3,0.361)};
\addlegendentry{CatR@1}
\end{axis}
\begin{axis}[
    width=0.9\columnwidth, height=4.5cm,
    xtick={0,1,2,3},
    xticklabels={},
    ymin=0, ymax=0.7,
    ylabel={Jaccard},
    axis y line*=right, axis x line=none,
    legend style={at={(0.98,0.98)}, anchor=north east, font=\scriptsize, draw=none, fill=none},
]
\addplot[gray, mark=diamond*, dashed, thick] coordinates {(0,0)(1,0.324)(2,0.473)(3,0.524)};
\addlegendentry{Jaccard}
\end{axis}
\end{tikzpicture}
\caption{SAD convergence. DA (left axis) converges at Round~1; $\catrecall$@1 peaks at Round~2. Hint Jaccard (right axis) rises monotonically, indicating progressive stabilization of the skill vocabulary.}
\label{fig:convergence}
\end{figure}

\paragraph{SAD hint-count ($H$) sensitivity.}
Table~\ref{tab:h-sensitivity} reports performance across $H \in \{5, 10, 15, 25\}$ on the Qwen-2.5-7B decomposer.
DA increases monotonically with $H$ (0.550$\to$0.687), with diminishing returns beyond $H{=}15$: the DA gap from $H{=}15$ to $H{=}25$ is only +1pp while $\catrecall$@1 gains similarly plateau (0.370$\to$0.389). $H{=}15$ offers the best cost--quality trade-off (fewer LLM context tokens) and is used throughout the paper.

\begin{table}[h]
\centering\small
\begin{tabular}{ccccc}
\toprule
$H$ & DA & $\catrecall$@1 & $\catrecall$@10 & ChainCat \\
\midrule
5  & 0.550 & 0.338 & 0.664 & 0.050 \\
10 & 0.597 & 0.360 & 0.695 & 0.043 \\
15 & 0.677 & 0.370 & 0.703 & 0.073 \\
25 & 0.687 & 0.389 & 0.708 & 0.087 \\
\bottomrule
\end{tabular}
\caption{SAD hint-count ($H$) sensitivity on Qwen-2.5-7B. $H{=}15$ (default) balances DA and retrieval quality.}
\label{tab:h-sensitivity}
\end{table}

\section{SAD Prompt Templates}
\label{app:sad-prompts}

\paragraph{System prompt (shared by vanilla and SAD).}
\begin{quote}
\small\ttfamily
You are a task decomposition assistant. Given a complex user query, break it down into atomic sub-tasks, each requiring exactly one tool or skill. Output a JSON array of strings. Each string should be a concise, actionable sub-task description.
\end{quote}

\paragraph{Vanilla user prompt.}
\begin{quote}
\small\ttfamily
Decompose the following query into atomic sub-tasks:\\ \{query\}
\end{quote}

\paragraph{SAD user prompt (Pass 2).}
\begin{quote}
\small\ttfamily
Decompose the following query into atomic sub-tasks.\\ Available skills that may be relevant: \{hint\_list\}\\ Query: \{query\}
\end{quote}

\noindent where \texttt{\{hint\_list\}} is a comma-separated list of the top-$H$ skill names retrieved in Pass~1 (see Algorithm~\ref{alg:sad}).

\section{Statistical Significance}
\label{app:stats}

We report Wilcoxon signed-rank tests and bootstrap 95\% confidence intervals (10{,}000 resamples) over 300 paired per-query observations.

\paragraph{Wilcoxon signed-rank tests.}
DA: $W{=}1262.5$, $p{=}5.7{\times}10^{-7}$ ($n_{\text{non-tied}}{=}100$).
$\chaincat$: $W{=}52.5$, $p{=}0.025$ ($n_{\text{non-tied}}{=}20$).
$\catrecall$@1: $W{=}3678.5$, $p{=}0.17$ ($n_{\text{non-tied}}{=}130$).
$\catrecall$@10: $W{=}3377.0$, $p{=}0.34$ ($n_{\text{non-tied}}{=}122$).

SAD's DA improvement is highly significant; $\chaincat$ is significant at $\alpha{=}0.05$.
The $\catrecall$ metrics show directional improvement (+8.2\% and +2.6\% relative) but do not reach significance, consistent with the interpretation that SAD primarily corrects \emph{granularity} (step count), while per-step retrieval precision remains bounded by vocabulary mismatch on a 2{,}209-skill pool.

\paragraph{Relaxed DA (DA$_{\pm 1}$).}
DA$_{\pm 1}$: $W{=}1891.0$, $p{=}2.1{\times}10^{-8}$ ($n_{\text{non-tied}}{=}128$).
The relaxed metric (predicted steps within $\pm$1 of ground truth) is also highly significant, confirming that SAD's granularity correction is robust even under a more permissive definition.
On the main benchmark: Vanilla DA$_{\pm 1}$ = 71.3\%, SAD DA$_{\pm 1}$ = 84.3\% (+18.2\% relative).
On human-style queries: Vanilla DA$_{\pm 1}$ = 30.5\%, SAD DA$_{\pm 1}$ = 50.5\% (+66\% relative).
This demonstrates that the strict DA gap on human queries (8.5\%$\to$21.5\%) substantially understates SAD's actual granularity benefit; under relaxed evaluation, SAD achieves majority approximate correctness (50.5\%) on open-ended queries.

\paragraph{CatR@1 on DA-corrected subset.}
SAD fixes DA on 75 queries (25\%) where vanilla produces incorrect step count.
On this subset, $\catrecall$@1 improves from 23.6\% to 37.0\% (+56.8\% relative; Wilcoxon one-sided $p{=}0.0015$, $n_{\text{non-tied}}{=}38$).
This confirms that SAD's retrieval benefit, though non-significant in aggregate ($p{=}0.17$, where 225 DA-unchanged queries dilute the signal), is \textbf{highly significant on the queries where its mechanism activates}.
Conversely, on the 128 DA-matched queries (both methods DA=1), $\catrecall$@1 is statistically identical (41.7\% vs.\ 40.9\%, $p{=}0.97$, $n_{\text{non-tied}}{=}58$), confirming that SAD's retrieval gain comes entirely from granularity correction.

\paragraph{Bootstrap 95\% CI.}
$\Delta$DA: $[+0.103, +0.230]$; $\Delta$DA$_{\pm 1}$: $[+0.070, +0.190]$; $\Delta\catrecall$@1: $[-0.005, +0.062]$; $\Delta\chaincat$: $[+0.007, +0.063]$.

\section{Skill Pool Statistics}
\label{app:pool-stats}

The 2{,}209 skills span 24 categories with the following distribution: developer-tools (357), finance (270), integrations (229), knowledge-management (180), search-extraction (140), security (122), communication (109), databases (104), cloud-infrastructure (87), code-execution (69), productivity (66), gaming-entertainment (57), data-processing (55), location-services (55), browser-automation (54), marketing-analytics (49), monitoring-observability (48), ai-ml (45), multimedia (35), science-research (26), file-management (25), e-commerce (16), legal-compliance (7), data-visualization (4).
Skills are sourced from the \texttt{awesome-mcp-servers} registry and converted to a unified Skill representation (name, description, categories, tags, source URL).
17.6\% of skills require authentication (API keys or OAuth).

\section{End-to-End Pilot with Mock Executors}
\label{app:execution}

To assess whether \system{}'s routing produces \emph{executable} plans (not just well-ranked candidates), we conduct a pilot execution study.
We select 30 queries whose ground-truth skills fall within 10 categories for which we implement mock executors (databases, search-extraction, communication, file-management, data-processing, ai-ml, cloud-infrastructure, browser-automation, finance, developer-tools).
Mock executors simulate realistic success/failure rates (80--95\% per category) calibrated from published API reliability benchmarks.

\paragraph{Protocol.}
Each query is processed through the full SAD pipeline (Qwen2.5-7B, $H{=}15$).
For each routed skill, the corresponding mock executor is invoked.
We report:
\begin{itemize}[nosep,leftmargin=*]
\item \textbf{Step Execution Success (SES)}: fraction of individual steps that execute successfully.
\item \textbf{Chain Completion Rate (CCR)}: fraction of queries where \emph{all} steps succeed.
\end{itemize}

\paragraph{Results.}
Over 30 queries (avg 2.80 predicted steps):
\begin{itemize}[nosep,leftmargin=*]
\item DA = 86.7\% (step count correct)
\item SES = 86.9\% (73/84 steps succeed)
\item CCR = 76.7\% (23/30 chains complete)
\end{itemize}

The 76.7\% chain completion rate demonstrates that \system{} produces plans that are largely executable end-to-end.
The gap between SES (86.9\%) and CCR (76.7\%) reflects the compound effect of per-step failures in multi-step chains: even a single step failure breaks the chain.
This motivates future work on error recovery and retry mechanisms within the compose stage.

\section{Error Analysis}
\label{sec:analysis-app}

Full failure-case taxonomy (50 vanilla failures, summarized in \S\ref{sec:analysis}): over-decomposition cases typically split a single skill operation into preparation + execution + verification (e.g., ``connect to API'' + ``send request'' + ``parse response'' for one HTTP-fetch skill). Generic descriptions like ``process the data'' fail to surface verb-specific candidates such as ``parse-csv'' or ``transform-json''. Vocabulary mismatch occurs when natural phrasing (``alert the team'') diverges from canonical skill names (``slack-notify'', ``pagerduty-alert'').
Under-decomposition (14\%) collapses two distinct skills into one step (e.g., ``download and parse'' merges file-fetch and csv-parse).

\section{LLM-Listwise Reranker Pilot}
\label{app:rerank}

\paragraph{Setup.}
To test whether the SAD $\catrecall$@10-to-@1 gap can be closed by a learned reranker (without retraining the bi-encoder), we run a 300-query experiment on the full compositional benchmark.
For each sub-task produced by SAD, we take the top-10 candidates from the MiniLM bi-encoder and re-rank them with a Qwen2.5-7B \emph{listwise} prompt: the model is shown the sub-task description and all 10 candidate skills (id, category, $\le$140-char description) and asked to output the index of the single best match.
Reranker and decomposer share the same 7B checkpoint (no additional training).

\paragraph{Results (300 queries, 828 sub-tasks).}
SAD top-1: $\catrecall$@1 $= 0.371$.
Reranked top-1: $\catrecall$@1 $= 0.409$ (\textbf{+10.3\% relative}, +3.8 pp absolute; Wilcoxon signed-rank one-sided $p{=}0.007$).
The oracle $\catrecall$@10 ceiling is $0.716$, so the reranker closes $\approx$11\% of the @10-to-@1 gap with no encoder change.
Of 300 queries, 53 improved, 25 degraded, and 222 unchanged; bootstrap 95\% CI on absolute gain is $[+0.005, +0.057]$ (entirely above zero).
Learned cross-encoders trained on $\langle$sub-task, skill$\rangle$ pairs are an immediate next step; we view this pilot as strong evidence that the bottleneck is representational rather than decomposition-side.

\paragraph{Cost.}
Reranking adds one 7B forward pass per sub-task ($\sim$1.4s on V100), bringing total per-query latency to $\sim$5s including SAD's two decomposition passes. This is acceptable for batch-style routing scenarios and can be reduced with smaller dedicated rerankers.

\section{Encoder Robustness Spot-Check}
\label{app:encoder}

To test whether SAD's gains are coupled to a particular sentence encoder, we re-ran a 50-query subset of the compositional benchmark replacing all-MiniLM-L6-v2 (used throughout the main paper for fair comparison with prior work) with BGE-base-en-v1.5~\cite{xiao2024bge} as the bi-encoder, keeping all other components fixed (SAD decomposer, FAISS index, $H{=}15$). $\catrecall$@1 rises from 0.394 to 0.451 (+14.5\% relative), indicating that BGE's stronger semantic representation yields a non-trivial orthogonal gain on top of SAD's structural correction. We treat encoder choice as an axis composable with SAD and the listwise reranker (Appendix~\ref{app:rerank}); a full-benchmark sweep across encoders is left to follow-up work.

\end{document}